\documentclass{article}

\usepackage{PRIMEarxiv}
\usepackage[utf8]{inputenc} 
\usepackage[T1]{fontenc}    
\usepackage{hyperref}       
\usepackage{url}            
\usepackage{booktabs}       
\usepackage{amsfonts}       
\usepackage{nicefrac}       
\usepackage{microtype}      
\usepackage{lipsum}
\usepackage{fancyhdr}       
\usepackage{graphicx}       
\usepackage{cite}
\usepackage{amsmath,amssymb,amsfonts}
\usepackage{algorithmic}
\usepackage{graphicx}
\usepackage{textcomp}
\usepackage{float}
\usepackage{algorithm}
\usepackage{array}
\usepackage{soul}
\usepackage{color, colortbl}
\usepackage{calc}

\title{Transfer learning-assisted inverse modeling in nanophotonics based on mixture density networks
}

\author{
  Liang Cheng \\
  Department of Information Technology \\
  Uppsala University \\
  SE 751 05, Uppsala, Sweden\\
  \And
  Prashant Singh \\
  Department of Information Technology \\
  Science for Life Laboratory, Uppsala University \\
  SE 751 05, Uppsala, Sweden\\
   \And
  Francesco Ferranti \\
  Department of Applied Physics and Photonics \\
  Vrije Universiteit Brussel and Flanders Make \\
  Pleinlaan 2, B-1050 Brussels, Belgium\\
  \texttt{francesco.ferranti@vub.be} \\
}

\begin{document}
\maketitle

\begin{abstract}

The simulation of nanophotonic structures relies on electromagnetic solvers, which play a crucial role in understanding their behavior. However, these solvers often come with a significant computational cost, making their application in design tasks, such as optimization, impractical. To address this challenge, machine learning techniques have been explored for accurate and efficient modeling and design of photonic devices. Deep neural networks, in particular, have gained considerable attention in this field. They can be used to create both forward and inverse models. An inverse modeling approach avoids the need for coupling a forward model with an optimizer and directly
performs the prediction of the optimal design parameters values.

In this paper, we propose an inverse modeling method for nanophotonic structures, based on a mixture density network model enhanced by transfer learning. Mixture density networks can predict multiple possible solutions
at a time including their respective importance as Gaussian
distributions. However, multiple challenges exist for mixture density network models. An important challenge is that an upper bound on the
number of possible simultaneous solutions needs to be
specified in advance. Also, another challenge
is that the model parameters must be jointly
optimized, which can result computationally expensive.
Moreover, optimizing all parameters simultaneously can
result numerically unstable and
lead to degenerate predictions. The proposed approach copes with these limitations using transfer learning-based techniques, while obtaining accurate results in the prediction of the design solutions given an optical response
as an input. A dimensionality reduction step is also explored. Numerical results validate the proposed method.
\end{abstract}

\section{Introduction}

Electromagnetic (EM) methods have become essential tools for simulating diverse and complex nanophotonic structures. However, EM simulations often entail a considerable computational cost. Design tasks, as optimization, involve multiple simulations for varying design parameter values. Directly employing EM solvers in the design workflow can be impractical due to the prohibitive computational cost associated with these simulations. Therefore, machine learning (ML) techniques have been explored for accurate and efficient modeling and design of photonic devices in the perspective of addressing this challenge. However, for the devices with complex expected targets, such as
the spectrum with multiple peaks and valleys, there are still many sufferings remaining for these
data-driven approaches, such as overfitting.

Photonics, a multidisciplinary field that encompasses the study and manipulation of light, has witnessed profound transformations thanks to ML techniques. ML techniques have been proposed in the literature for the modeling and design optimization of photonic devices \cite{Ma2018,Peurifoy2018,Liu2018,Zheng20,Wiecha20,Wiecha21,So2021,Roberts21,Unni2021,Kojima2021,Xu2021,Wiecha22,Ferranti22_2,Padilla22,Yesilyurt23,Luce2023}. Deep neural networks (DNNs) are frequently employed, utilizing functions constructed by combining multiple layers of affine transformations and non-linear activation functions, which allows modeling intricate patterns in a data-centric manner. 
DNNs can be used to create both forward and inverse models. In the forward network model, the inputs comprise the design parameters (e.g., geometrical parameters), while the outputs represent the optical response. In an inverse model, the optical response is provided as an input, and the model predicts the design parameters values (outputs) that lead to that optical response. In the case of photonics modeling, the optical response is often represented by the wavelength-dependent response sampled at a set of wavelength values. Multiple techniques have been proposed for the inverse modeling \cite{Zheng20,Wiecha21,Padilla22,Yesilyurt23,Luce2023}. An inverse modeling approach avoids the need for coupling a forward model with an optimizer and directly performs the prediction of the optimal design parameters values.

This paper focuses on the Mixture Density Network (MDN) architecture for inverse modeling \cite{Zheng20,Unni2021,Yesilyurt23}, which is based on the composition of a Gaussian Mixture Model and a Neural Network. The output space of an MDN contains the mean and standard deviation values of multiple Gaussian probability distribution functions (pdfs) and the mixing coefficients that denote some weights (importance) associated with the Gaussian pdfs. MDNs can predict multiple possible solutions
at a time including their respective importance as Gaussian
pdfs. However, multiple challenges exist for MDN models. An important challenge is that an upper bound on the
number of possible simultaneous solutions needs to be
specified in advance. Another challenge
is that the model parameters must be jointly
optimized, which can result computationally expensive.
In addition,
an optimization of all parameters simultaneously can result
numerically unstable and lead to degenerate predictions. The
approach proposed in this paper copes with these limitations
by proposing transfer learning techniques, while obtaining
accurate results in the prediction of the design solutions given
an optical response as an input. A dimensionality reduction
step is also explored.

\section{Proposed inverse modeling techniques}
\subsection{Mixture Density Networks}
MDNs are built from two main components, a Gaussian Mixture Model and a Neural Network. The output space of an MDN comprises the mean, standard deviation and mixing coefficient values of multiple Gaussian pdfs. This allows handling the possible multi-value mapping problem that is often encountered in inverse modeling. This multi-value mapping problem arises from the fact that different values of the design parameters can correspond to very similar optical responses.
Given the input $\mathbf{x}$ (optical response) and output $\mathbf{y}$ (design space parameters), the pdf $\mathrm{p}(\mathbf{y} \mid \mathbf{x})$ can be expressed as a linear combination of pdfs:



\begin{align}
\mathrm{p}(\mathbf{y} \mid \mathbf{x})&=\sum_{k=1}^{K} \Pi_{k} (\mathbf{x})\phi_{k}(\mathbf{y} \mid \mathbf{x})=\sum_{k=1}^{K} \Pi_{k} (\mathbf{x})\phi_{k}(\mathbf{y}, \pmb{\mu}_{k}(\mathbf{x}),\pmb{\sigma}_{k}(\mathbf{x}))
\label{eq:total_pdf}
\end{align}

\noindent where the mixing coefficients $\Pi_{k} (\boldsymbol{x})$ can be regarded as importance weights and $\phi_{k}$ are the pdfs. The Gaussian pdf is used herein as a sufficient number of mixed Gaussian pdfs can approximate complex distributions:


\begin{align}
\phi_{k}(\mathbf{y}, \pmb{\mu}_{k}(\mathbf{x}), \pmb{\sigma}_{k}(\mathbf{x})) = \prod_{c=1}^{N} \left(\frac{1}{\sqrt{2 \pi}\sigma_{k,c}(\mathbf{x})} exp\left({-\left(\frac{y_c-\mu_{k,c}(\mathbf{x})}{\sqrt{2}\sigma_{k,c}(\mathbf{x})}\right)^2}\right)\right)
\end{align}



 \noindent where $N$ is the dimension of the output vector $\mathbf{y}$ (number of design parameters). 
The standard deviations $\pmb{\sigma}_{k}$ are positive, $\Pi_{k}$ values lie between $0$ and $1$ and $\sum_{k=1}^{K} \Pi_{k} = 1$.

When an MDN is applied on nanophotonics data, the input can be a high dimensional vector related to the sampled optical response as a function of the wavelength, which may increase the CPU time for the model training. A dimensionality reduction method may be used to accelerate the model training. An autoencoder is explored in this paper, which is a supervised neural network that learns to encode a high dimensional representation into a low dimensional representation (latent space) and then reconstruct the full representation by a decoder step. 


\subsection{Network Structure}

\begin{figure}[h]
\small
\centering
\includegraphics[height=3cm,width=9cm]{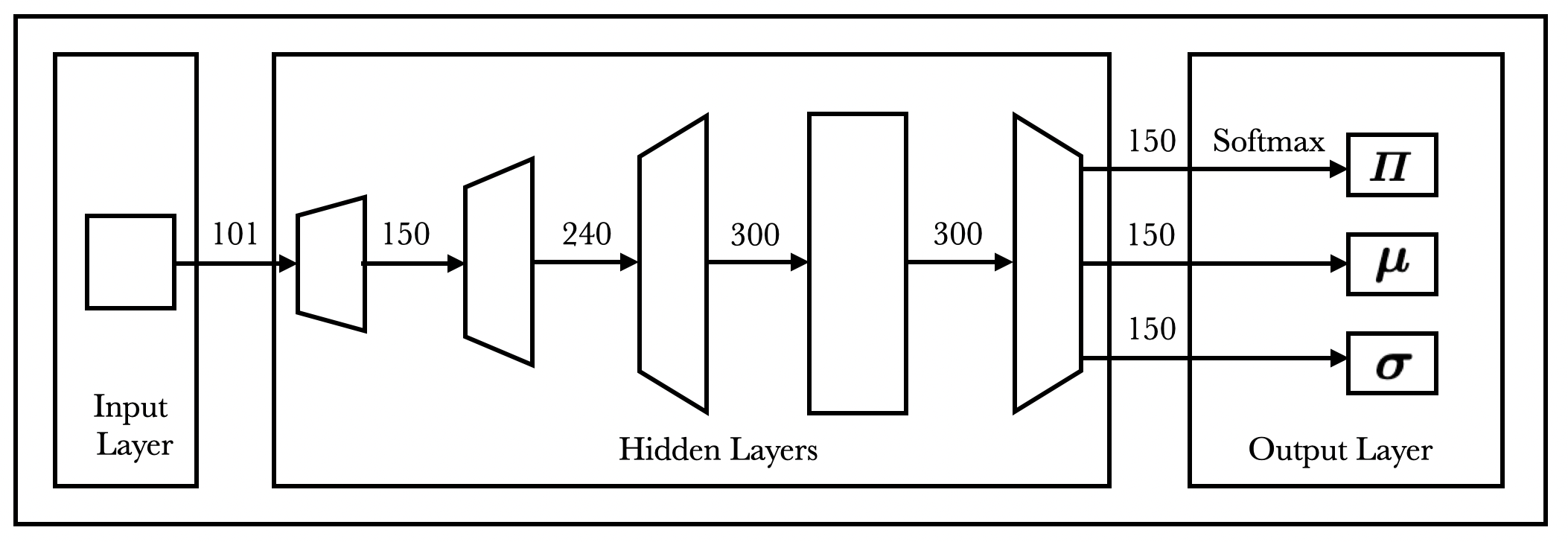}
\includegraphics[height=3cm,width=9cm]{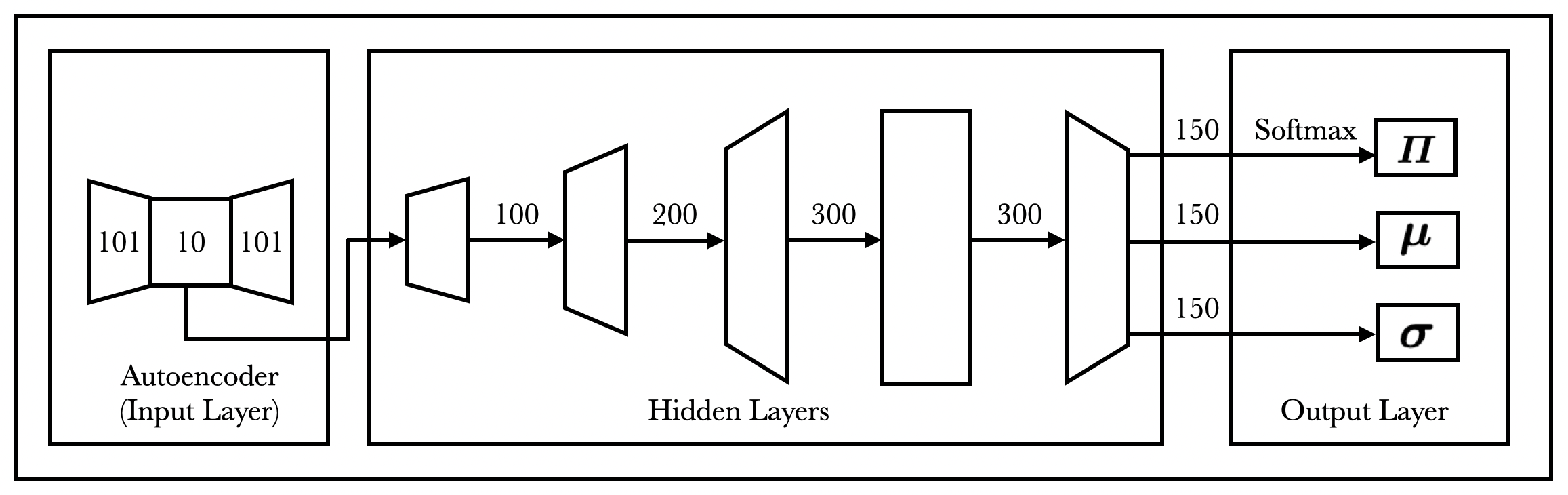}
\caption{MDN structure without and with an autoencoder.} \label{fig:MDN1}
\end{figure}

The structure of an MDN without (top) and with (bottom) an autoencoder (the numerical values in these figures are for illustration purposes) is shown in Fig. \ref{fig:MDN1}. The input data is a set of optical spectra discretized over the wavelength axis, and the hidden layers consist of fully-connected layers and activation functions. The first few hidden layers gradually expand the data size in order to allow the model to better capture the local patterns present in the input data. Once the size reaches the maximum, dropout is performed to improve the generalization ability of the model. After dropout, the data is downscaled to a reasonable size suitable for processing in the output layer. The output layer finally generates the key parameters $\pmb{\Pi}$, $\pmb{\mu}$ and $\pmb{\sigma}$ as the outputs of the MDN. 


The loss function used in an MDN is the negative log likelihood (NLL) defined as:


\begin{equation}
-\log(\mathrm{p}(\mathbf{y} \mid \mathbf{x}))=-\log \left[\sum_k^K \Pi_k(\mathbf{x}) \phi_k\left(\mathbf{y}, \pmb{\mu}_k(\mathbf{x}),\pmb{\sigma}_k(\mathbf{x})\right)\right]
\end{equation}

\noindent During model training in our numerical example, the NaN loss issue occurred multiple times, possibly
caused by the logarithm acting on a value too close to zero or by gradients becoming very small as they propagate through
multiple layers. A method proposed in \cite{She22} was used to cope with the NaN loss problem for the numerical example shown in what follows. It adds a small number like $10^{-5}$ in the loss function, and the updated loss function is defined as follows:

\begin{equation}
-\log \left[ \left( \sum_k^K \Pi_k(\mathbf{x}) \phi_k\left(\mathbf{y}, \pmb{\mu}_k(\mathbf{x}), \pmb{\sigma}_{k}(\mathbf{x})+10^{-5}\right)\right) +10^{-5}\right]
\label{eq:MDN_loss2}
\end{equation}

Other methods also exist to cope with the same problem and have better numerical stability, however an extensive discussion on this aspect is beyond the scope of this paper. This cost function was used consistently for all numerical results.

\subsection{Transfer Learning}

An important challenge in MDN is selecting the proper number of mixture components $K$. However, this upper bound is typically unknown necessitating exploration of multiple different mixture component numbers. Also, another challenge of training an MDN is that the model parameters of $\Pi$, $\pmb{\mu}$ and $\pmb{\sigma}$ must be jointly optimized,
which can result computationally expensive. Moreover, optimizing all parameters simultaneously can be numerically unstable and can lead to degenerate predictions. We explored transfer learning as a way to cope with these issues.

The proposed transfer learning approach starts with $K=1$ to fully train the inverse model. Thereafter, model training with the next mixture component number reuses trained parameters, effectively shortening the training time. For the MDN models with different mixture component numbers, the hidden layers structure would still be the same and the only difference in model structure is the output layer. Therefore, the parameters of the hidden layers can be initialized with the hidden layers parameters of the previous model. As for the parameters initialization in the output layer, we propose two different methods. 

The inverse model consists of three main outputs quantities, corresponding to $\pmb{\Pi}$, $\pmb{\mu}$ and $\pmb{\sigma}$, respectively. Based on the structure of the loss function, the $\pmb{\Pi}$ coefficients are an important factor in enabling the model to learn. In order to retain all the trained pdfs from the previous model and provide the newly added distribution equal opportunities to learn during the new round of training, the $\Pi_k$ terms can be initialized to be equal. Due to the nature of the softmax function used in the output layer of $\pmb{\Pi}$, initializing all the parameters for $\pmb{\Pi}$ to be zero is one option to achieve this
goal, since this provides $\Pi_k=1/K$. As for the output layer of $\pmb{\mu}$ and $\pmb{\sigma}$, two initialization schemes are proposed. In the first approach, the first $K-1$ distributions inherit the parameters from the previous model, while the newly added distribution will randomly select the parameters of one distribution from the $K-1$ distributions as its parameters. In the second approach, the first $K-1$ distributions also inherit the same parameters as the previous model, but the new distribution uses the parameters corresponding to the distribution associated with the first mixing coefficient in the previous model as its parameters instead of a random selection from the previous $K-1$ distributions. The first method is able to encourage the model to learn more diversified possible solutions with a randomness exploration characteristic. On the other hand, the second method can reduce the randomness in training, but may generally lead to a situation where most of the distributions are likely to be relatively similar to each other. Other transfer learning strategies can be used.

\section{Numerical results}\label{sec:results}

\begin{figure}[!ht]
    \centering
    \includegraphics[width=0.3\linewidth]{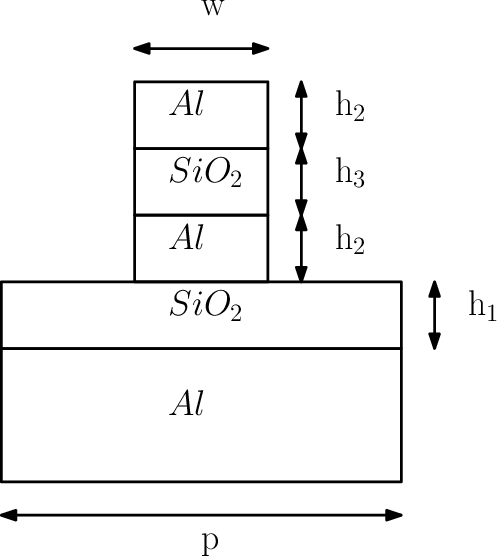}
    \caption{Cross-section of the unit cell.}
    \label{fig:Figure_MM_layout}
\end{figure}

A grating-based multiband absorber is the device of interest in this example. It is a periodic structure with a unit cell composed of a five-layered metal-insulator-metal configuration as illustrated in Fig. \ref{fig:Figure_MM_layout}. A periodic unit cell with an
EM wave at normal incidence on top of the unit cell was simulated electromagnetically using a frequency-domain solver of COMSOL Multiphysics\textsuperscript{\textregistered}. The wavelength ($\lambda$) interval is in the visible range $[400-700]$ nm.

The design parameters of the unit cell are: the unit cell period
$p$, the width $w$ and the thicknesses $h_1$, $h_2$, $h_3$. The intervals of variation are [305-415] nm, [45-190] nm, [150-295] nm, [25-200] nm, and [80-165] nm, respectively. $101$ wavelength samples are chosen over the wavelength interval [400-700] nm. $3848$ scattered samples in the design space are chosen based on a Sobol scrambled quasi-random sequence scheme \cite{Brandimarte14}. A possible fabrication constraint on the design parameters samples is considered, namely $p-w \geq 200$ nm, to ensure a certain space between unit cells. The EM absorbance response of this device is considered in this example.

All model training and tests were performed on a MacOS 13.4.1 platform consisting of the Apple M1 CPU (8 cores) and 16 GB RAM. The code was implemented in PyTorch. The loss function used for the training is described in (\ref{eq:MDN_loss2}). The total number of spectra used in this work is 3848, and these spectra are divided into training, validation and test datasets according to the ratio $80\%-10\%-10\%$. Early stopping, a measure to avoid model overfitting, is used in the training process. Table \ref{tab:neuralnetwork} provides information about the used model architecture ($K$ denotes the number of Gaussian pdfs and $5$ is the number of design parameters). 

\begin{table*}[h]
\centering
\small
\caption{MDN and AE models information.}
\begin{tabular}{|l|c|}
\hline \# neurons (MDN) & {$[101,150,240,300,300,150,[K,5K,5K]]$} \\
\# neurons (MDN+AE) & {$[10,100,200,300,300,150,[K,5K,5K]]$} \\
\# neurons (AE) & {$[101,128,256,512,256, 10,256,512,256,128,101]$} \\
Optimizer & Adam \\
Learning rate & $10^{-3}$ \\
Activation functions & Sigmoid Linear Unit (SiLU) \\
\hline
\end{tabular}\label{tab:neuralnetwork}
\end{table*}

\begin{figure}[h]
\small
\centering
\includegraphics[height=8cm,width=6cm,angle=-90]{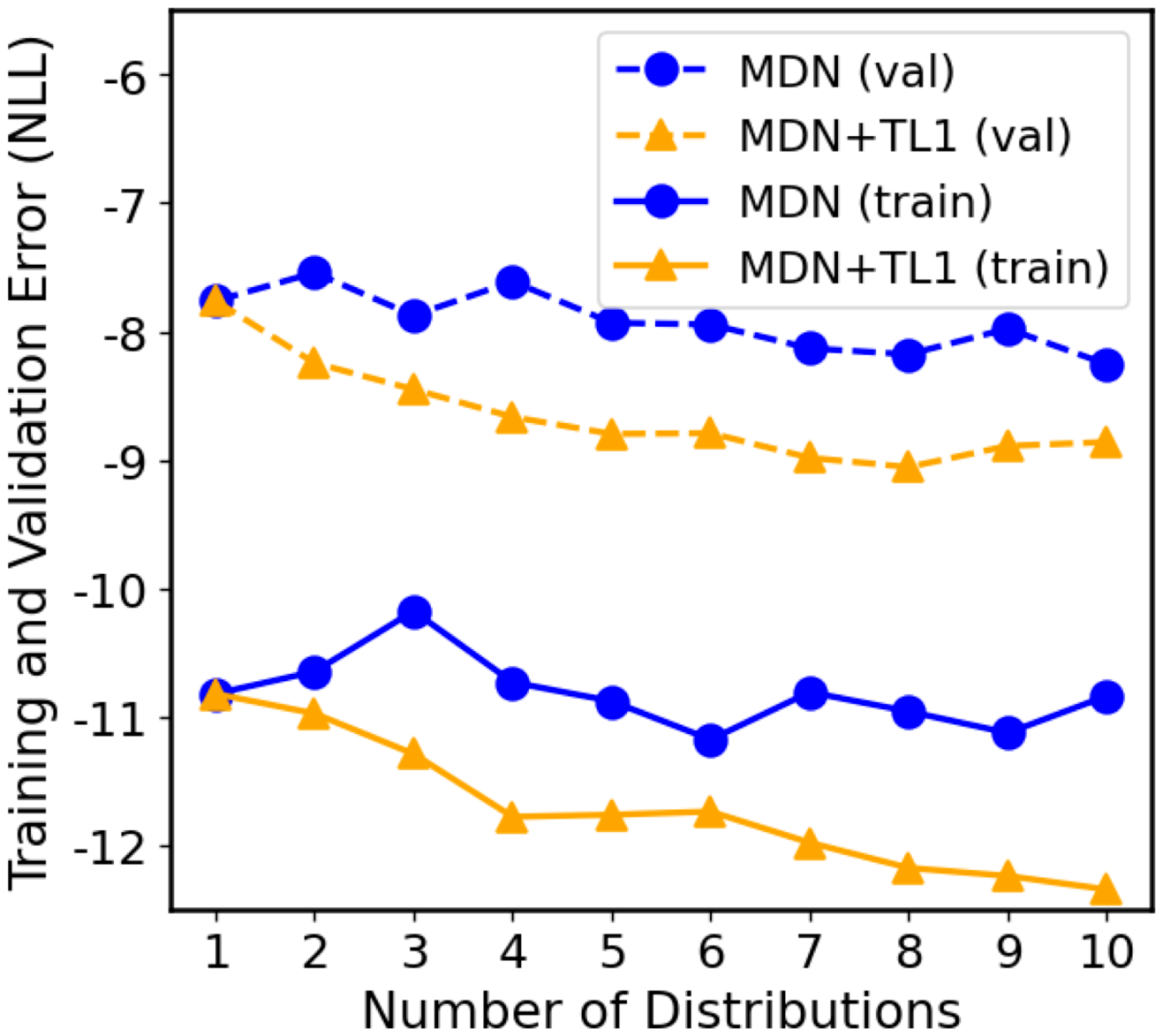}
\includegraphics[height=8cm,width=6cm,angle=-90]{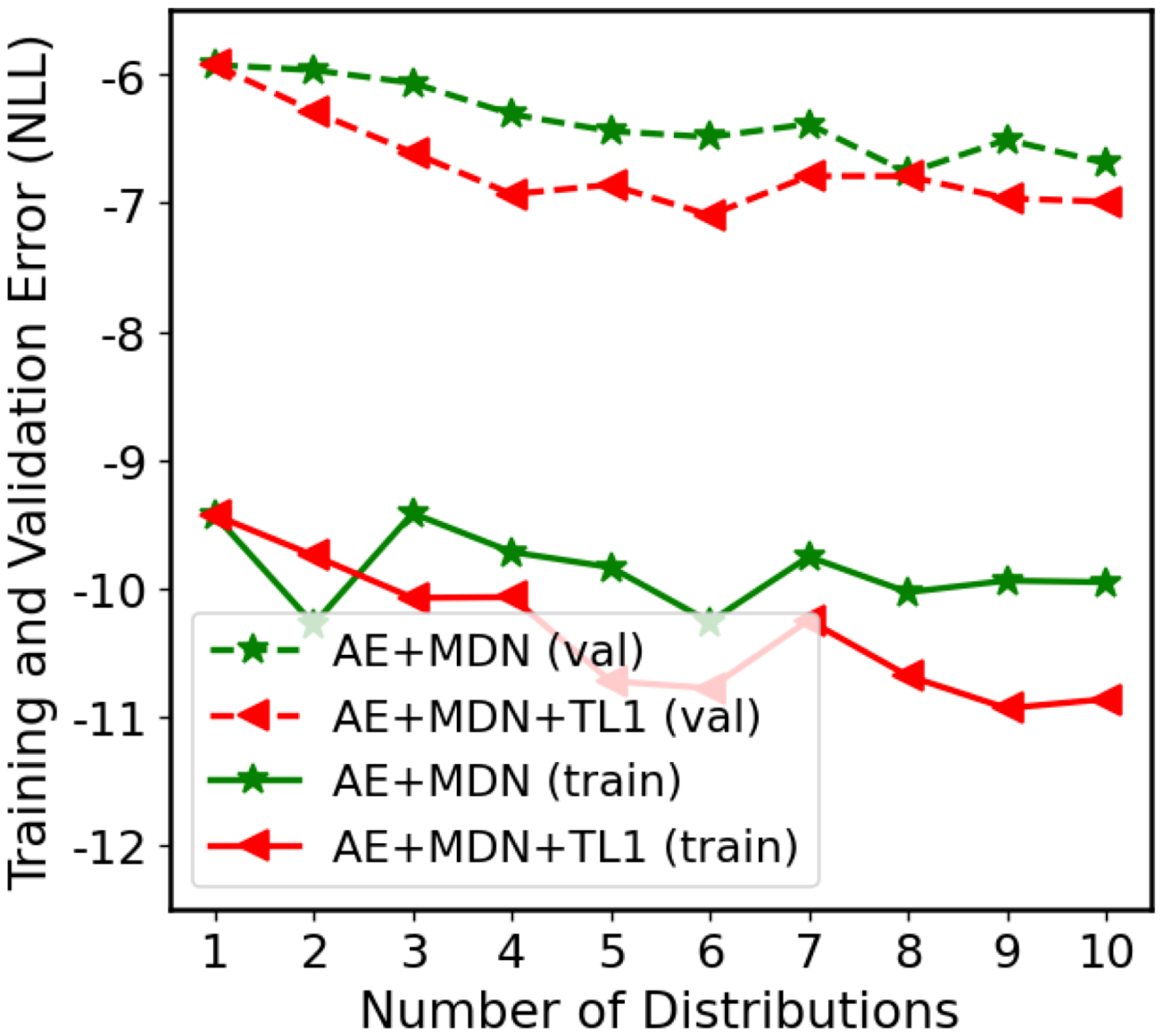}
\caption{Training and validation loss of MDNs and AE+MDNs with and without transfer learning (TL1).} \label{fig:loss}
\end{figure}

\begin{figure}[h]
\small
\centering
\includegraphics[height=8cm,width=6cm,angle=-90]{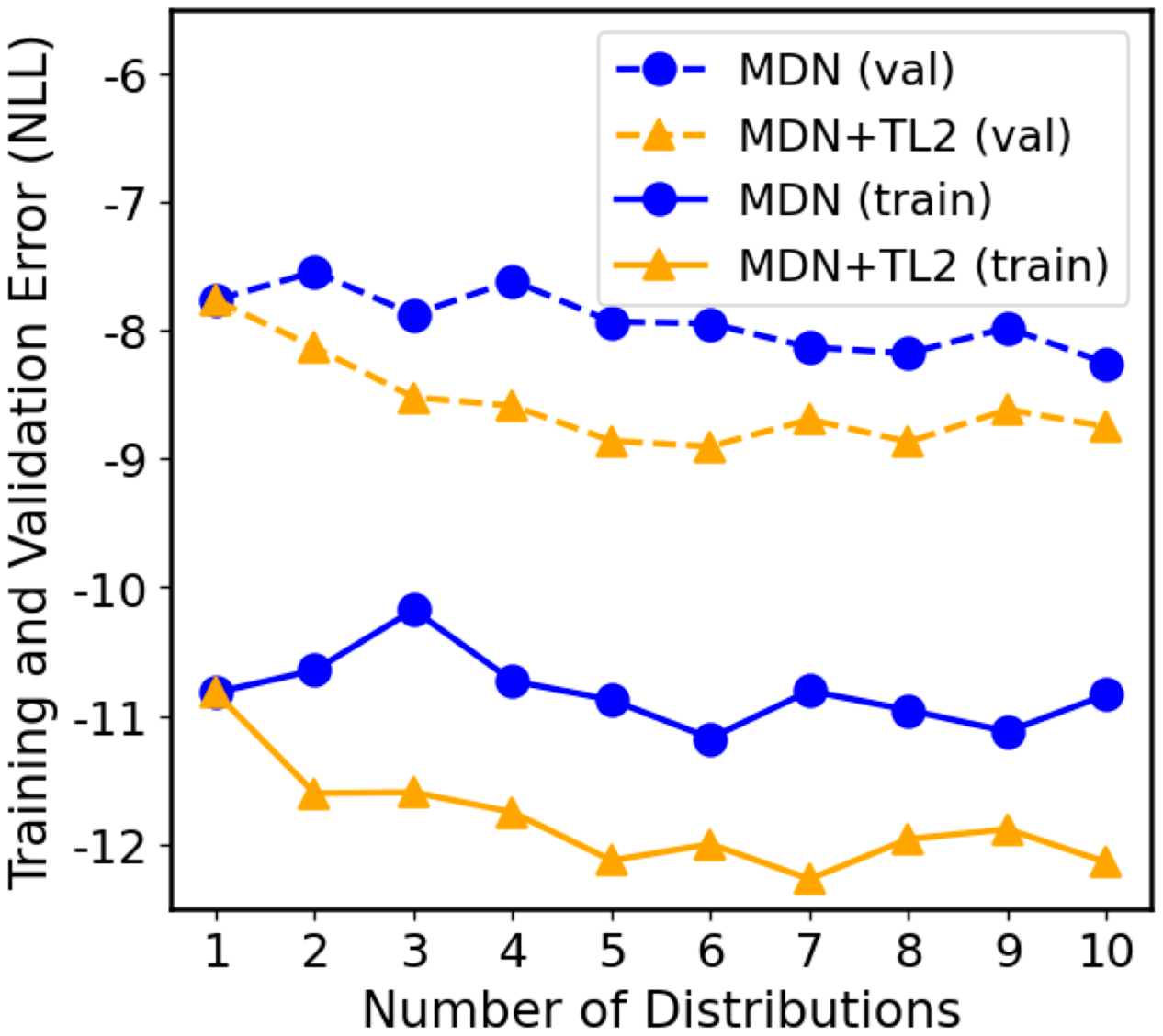}
\includegraphics[height=8cm,width=6cm,angle=-90]{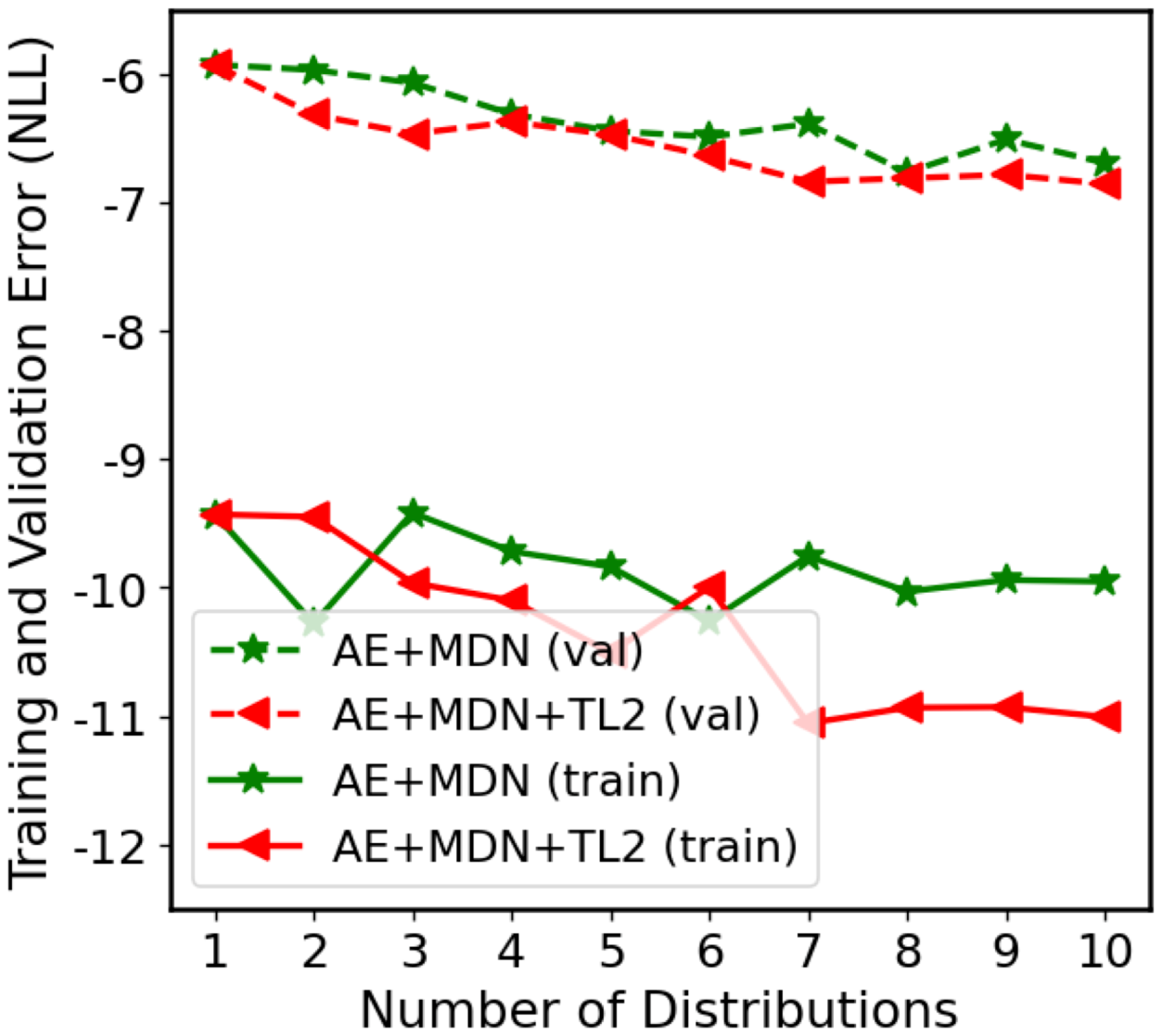}
\caption{Training and validation loss of MDNs and AE+MDNs with and without transfer learning (TL2).} \label{fig:loss2}
\end{figure}

\begin{figure}[h]
\small
\centering
\includegraphics[height=7cm,width=5cm,angle=90]{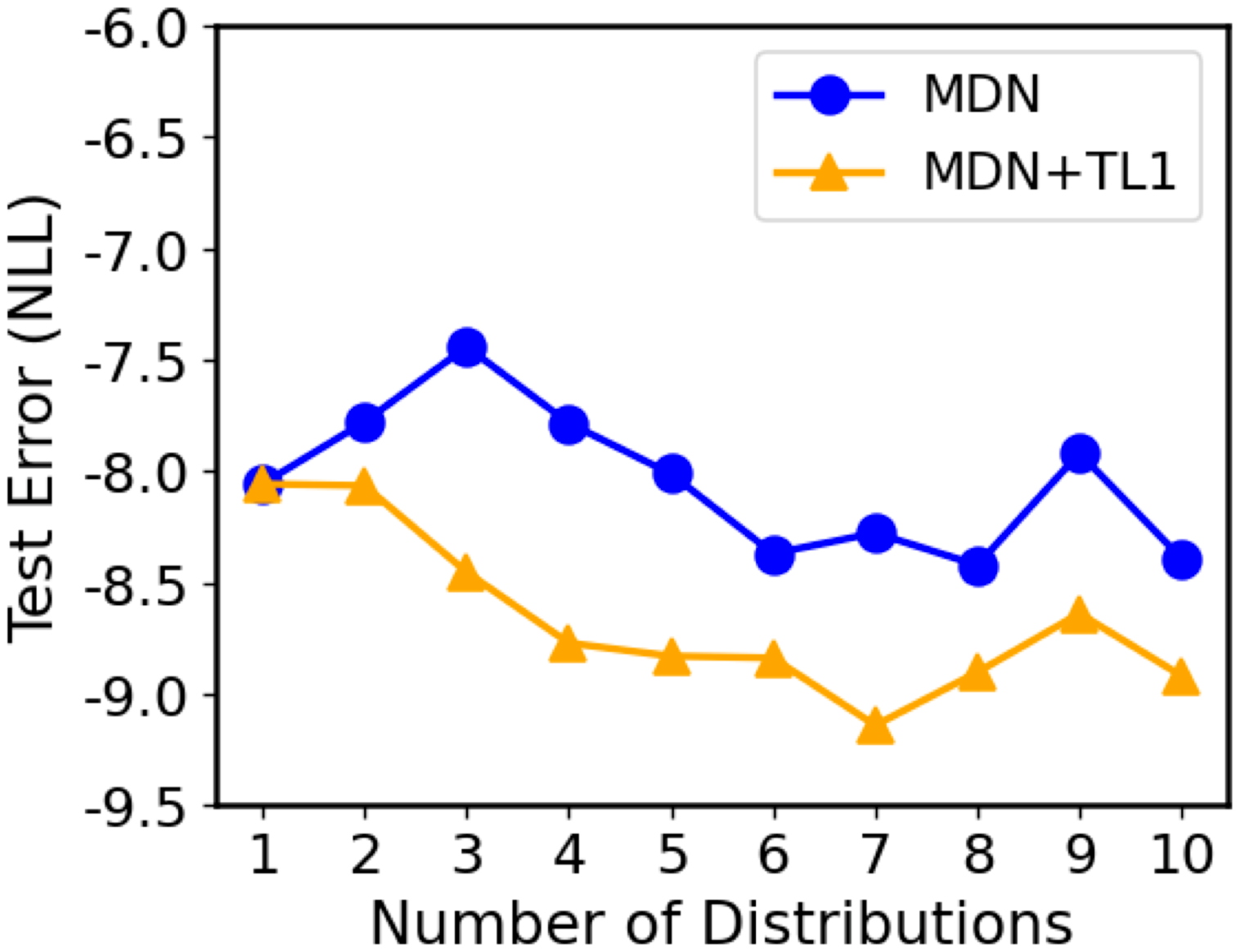}
\includegraphics[height=5cm,width=7cm,angle=0]{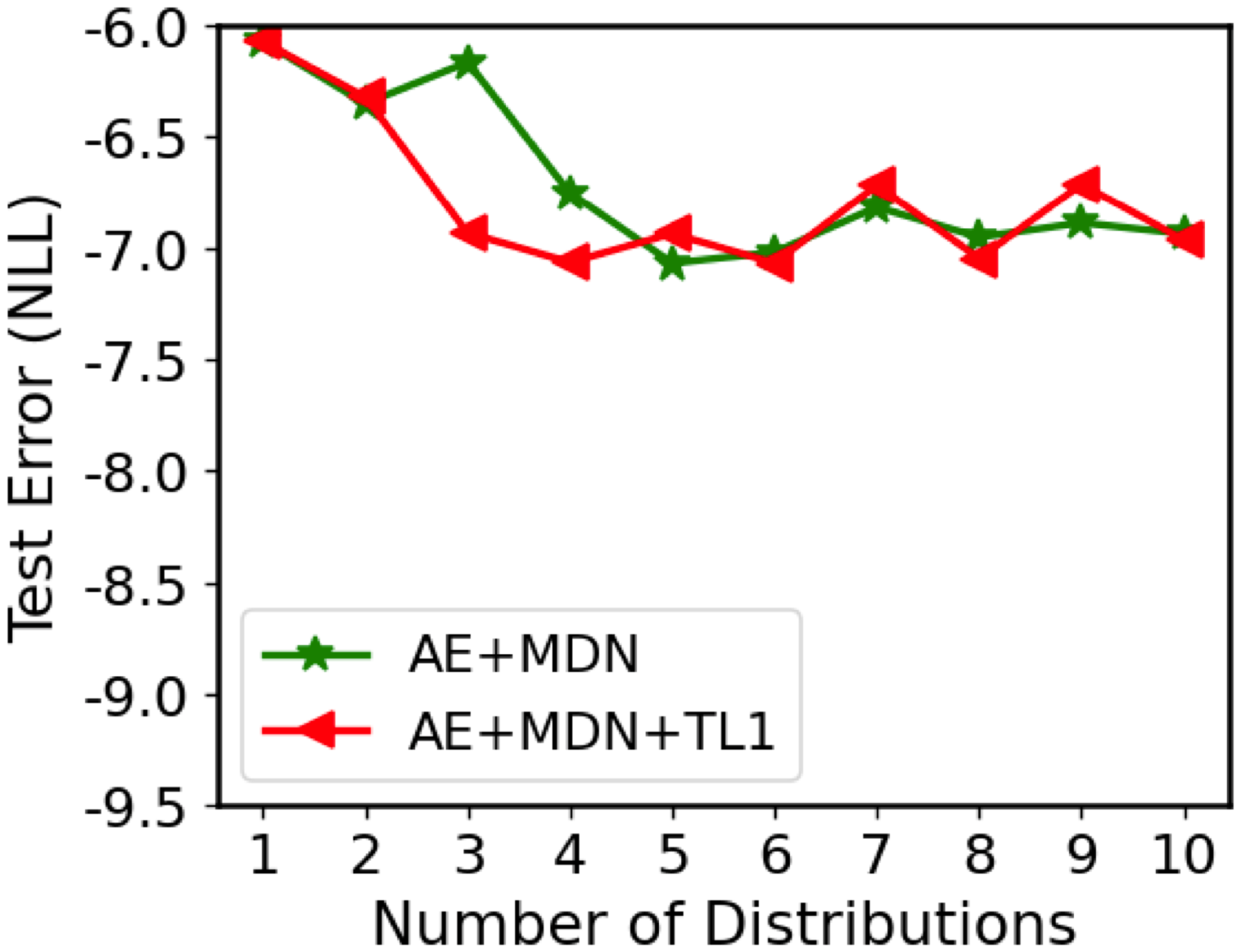}
\caption{Test loss of MDNs and AE+MDNs with and without transfer learning (TL1).} \label{fig:test_error_nll}
\end{figure}

\begin{figure}[h]
\small
\centering
\includegraphics[height=7cm,width=5cm,angle=90]{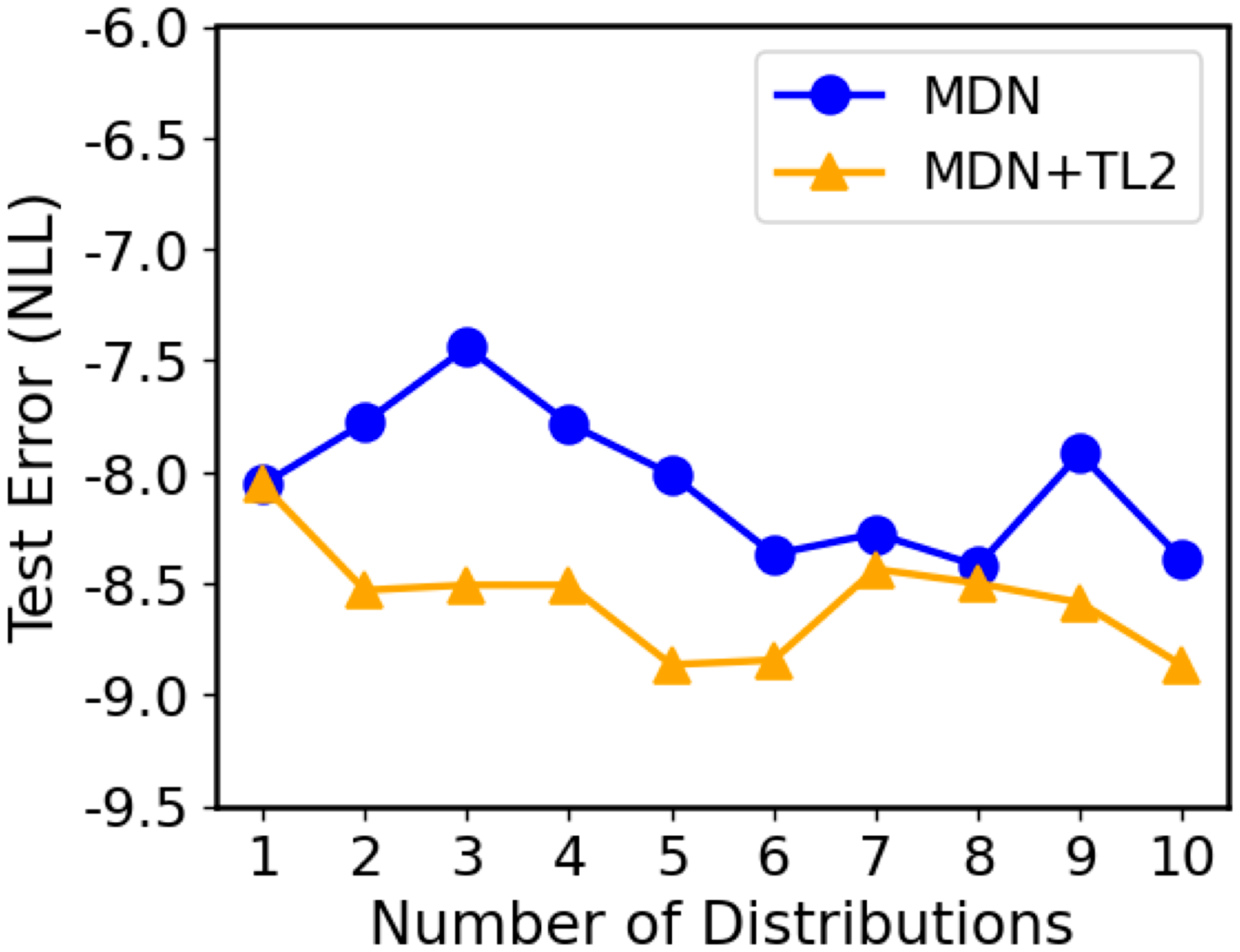}
\includegraphics[height=7cm,width=5cm,angle=90]{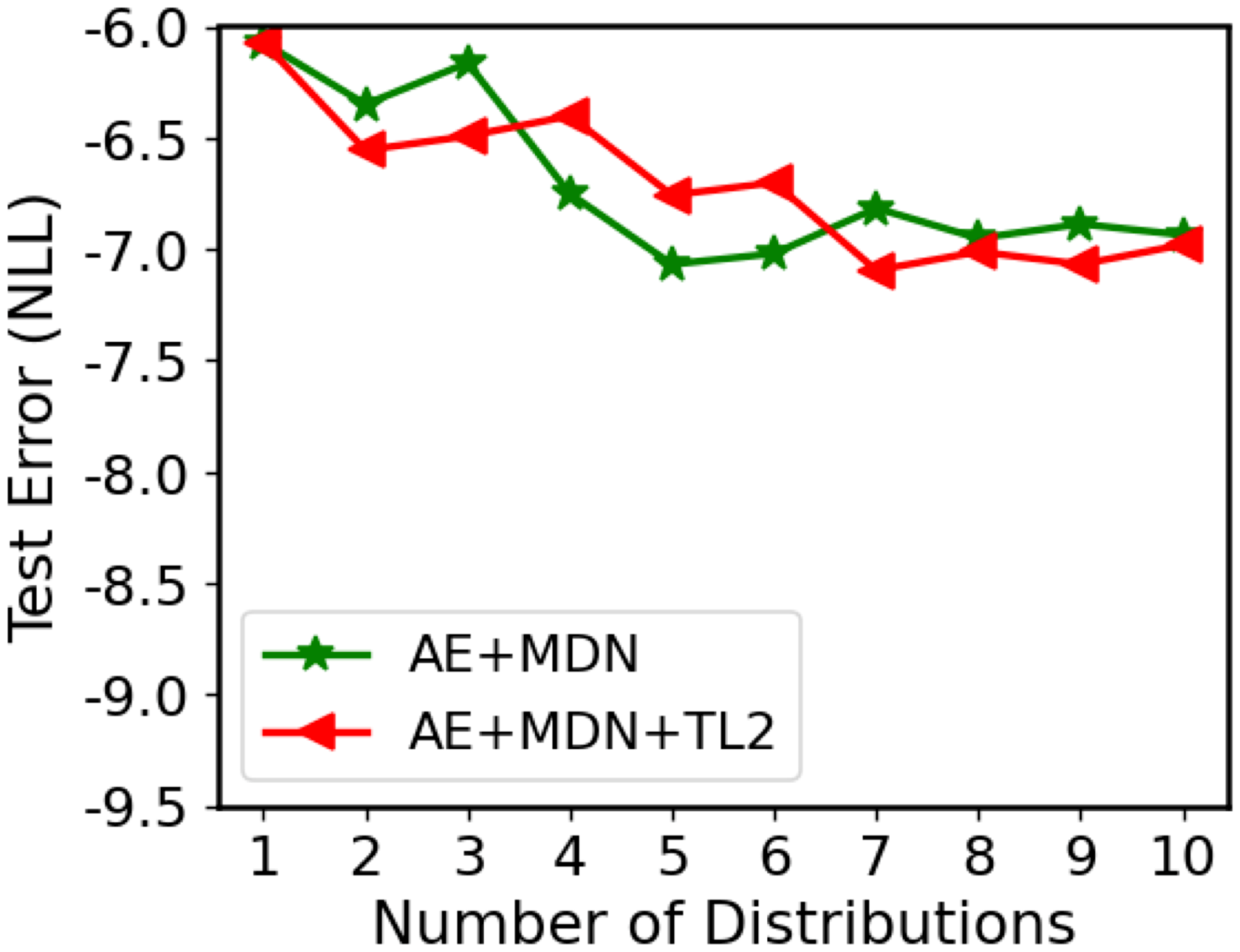}
\caption{Test loss of MDNs and AE+MDNs with and without transfer learning (TL2).} \label{fig:test_error_nll_2}
\end{figure}

A limit of 1000 epochs was set for training, subject to early-stopping. Table \ref{tab:cputime} summarizes the CPU time needed for the constructions of the MDNs ($K=1,...,10$) with and without autoencoder and with and without transfer learning approaches. TL1 and TL2 refer to the first and second proposed transfer learning strategies. When the autoencoder is used the CPU time to train the autoencoder is included. Figs. \ref{fig:loss}-\ref{fig:test_error_nll_2} provide information about the accuracy of all these models considering training, validation, and test errors based on the NLL cost function in (\ref{eq:MDN_loss2}). 

\begin{table*}[h]
\centering
\small
\caption{CPU time comparison for the MDNs with and without autoencoder (AE) and transfer learning (TL)}
\begin{tabular}{|c|c|c|c|c|c|c} 
\hline
No AE/TL & AE & TL1 & AE+TL1 & TL2 & AE+TL2\\
\hline $31 \mathrm{~min}$ & $28 \mathrm{~min}$ & $6 \mathrm{~min}$ & $10 \mathrm{~min}$ & $6 \mathrm{~min}$ & $10 \mathrm{~min}$\\ 
\hline
\end{tabular}\label{tab:cputime}
\end{table*}



These results show that the proposed transfer learning approaches work well and allow reducing the training CPU time needed for multiple MDNs with a different number of Gaussian pdfs, while obtaining accurate results. The autoencoder has its training computational cost, it helps to reduce the training CPU time only when a transfer learning approach is not used, however there is a degradation in the accuracy performance of the MDNs overall. 



\begin{figure}[!ht]
    \centering
    \includegraphics[height=8cm,width=9.5cm]{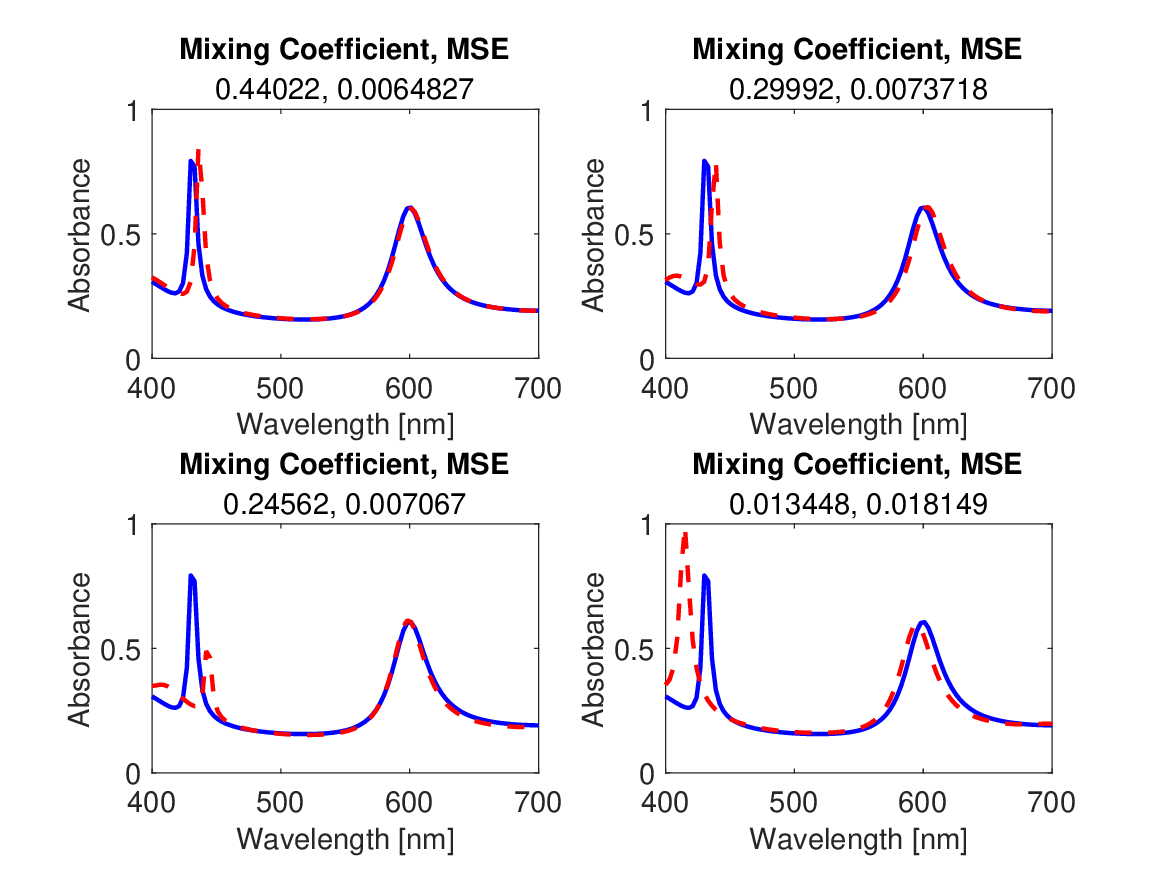}
   \caption{Prediction of the MDN model.}
    \label{fig:results1}
\end{figure}
\begin{figure}[!ht]
    \centering
     \includegraphics[height=8cm,width=9.5cm]{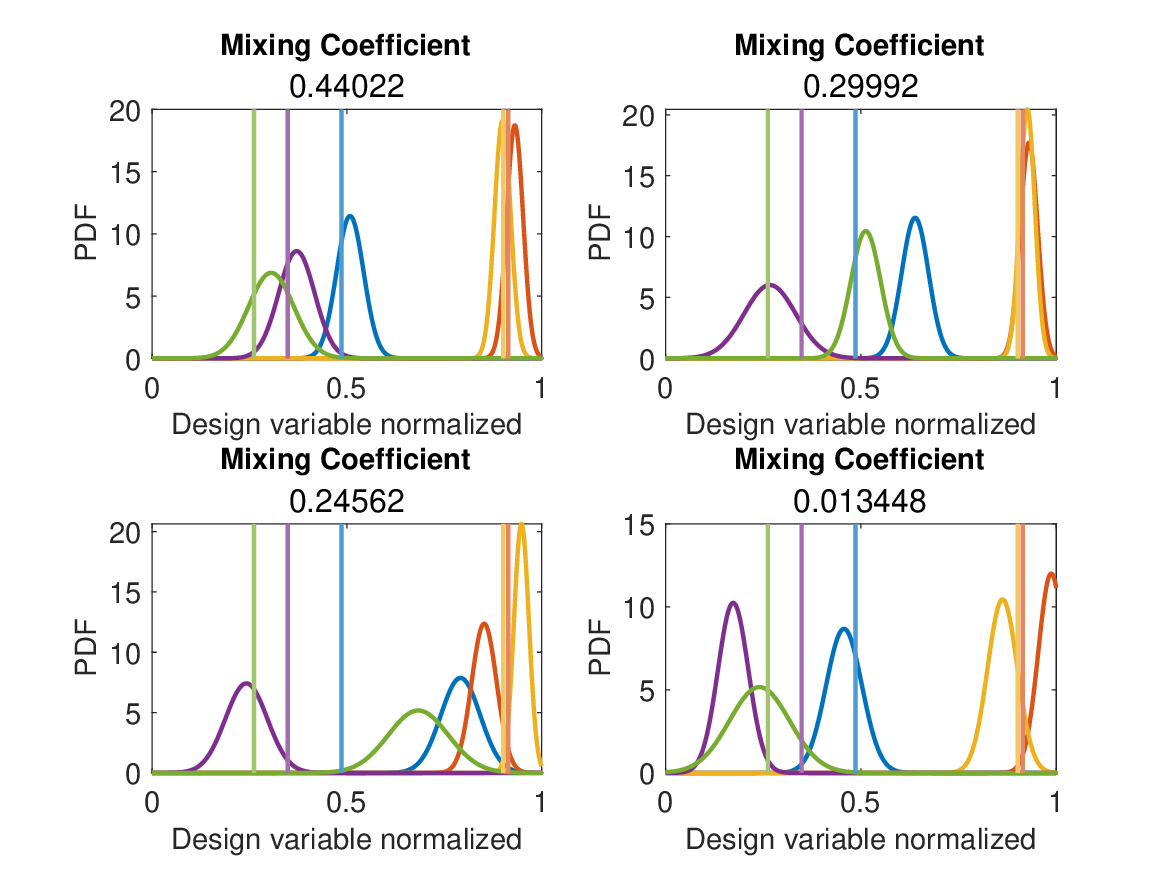}
   \caption{Gaussian pdfs of the MDN model.}
    \label{fig:results1_2}
\end{figure}
\begin{figure}[!ht]
    \centering
    \includegraphics[height=8cm,width=9.5cm]{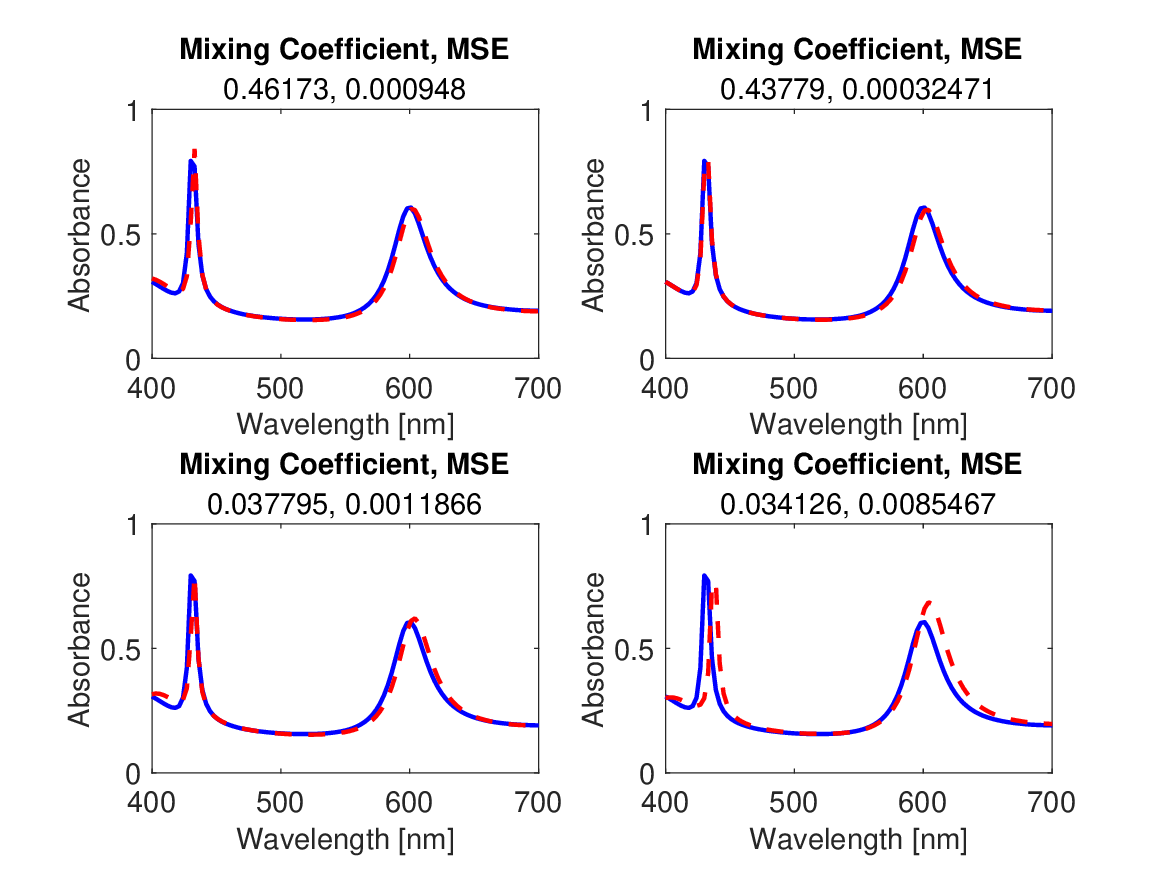}
   \caption{Prediction of the MDN+TL1 model.}
    \label{fig:results2}
\end{figure}
\begin{figure}[!ht]
    \centering
    \includegraphics[height=8cm,width=9.5cm]{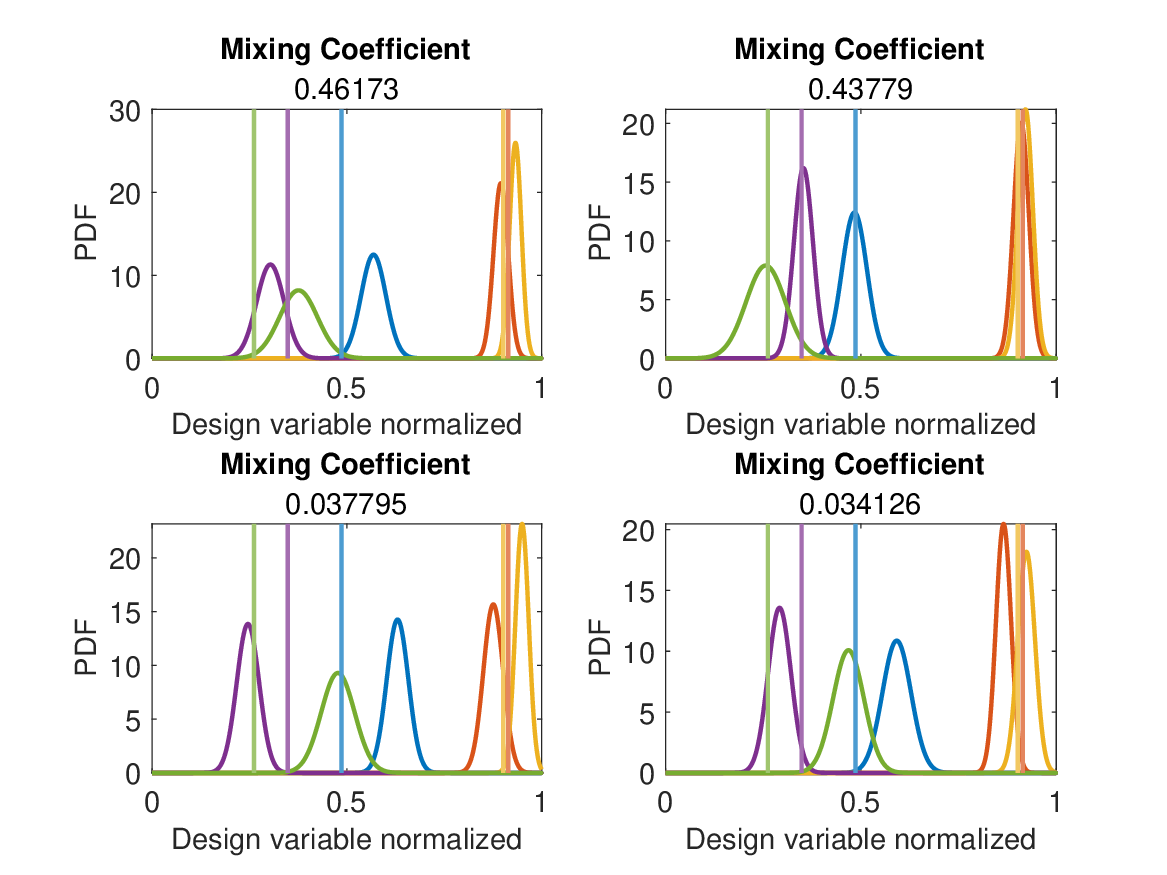}
   \caption{Gaussian pdfs of the MDN+TL1 model.}
    \label{fig:results2_2}
\end{figure}
\begin{figure}[!ht]
    \centering
   \includegraphics[height=3.7cm,width=4.7cm]{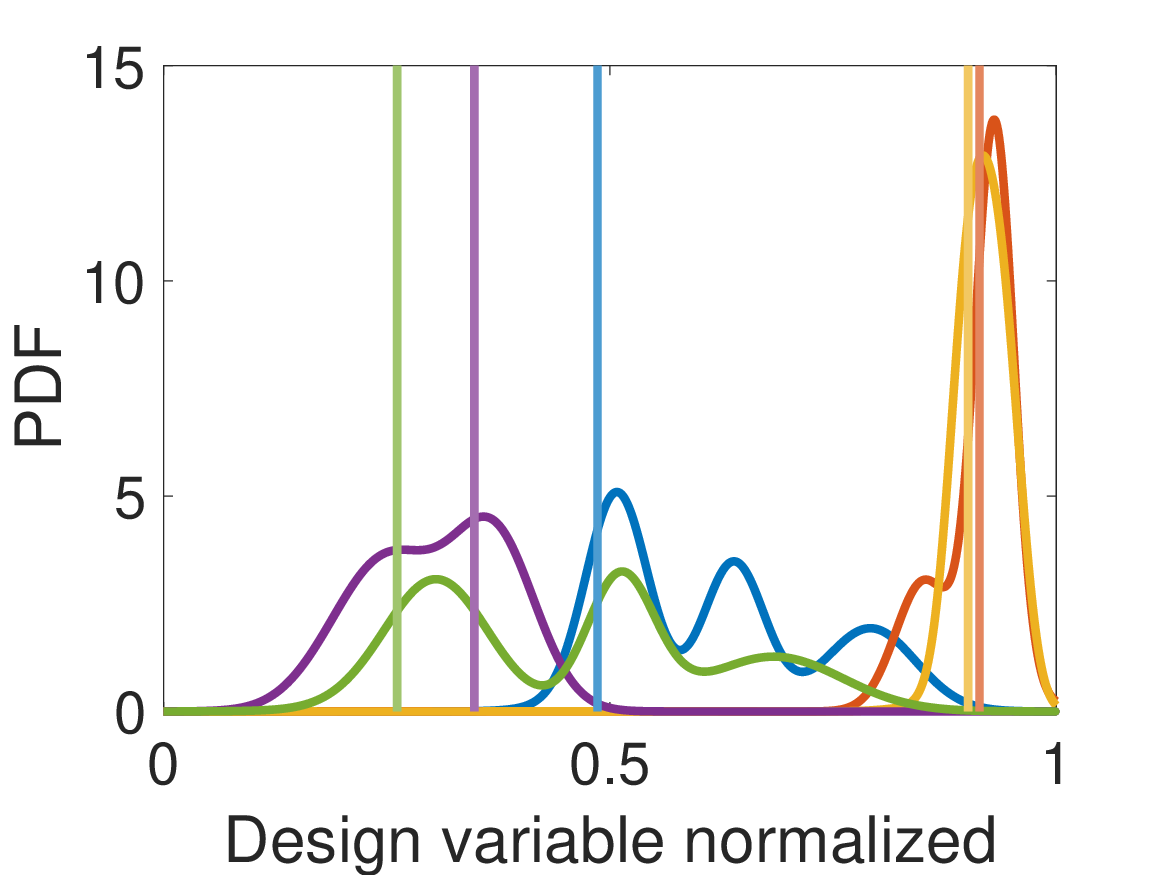}
 \includegraphics[height=3.7cm,width=4.7cm]{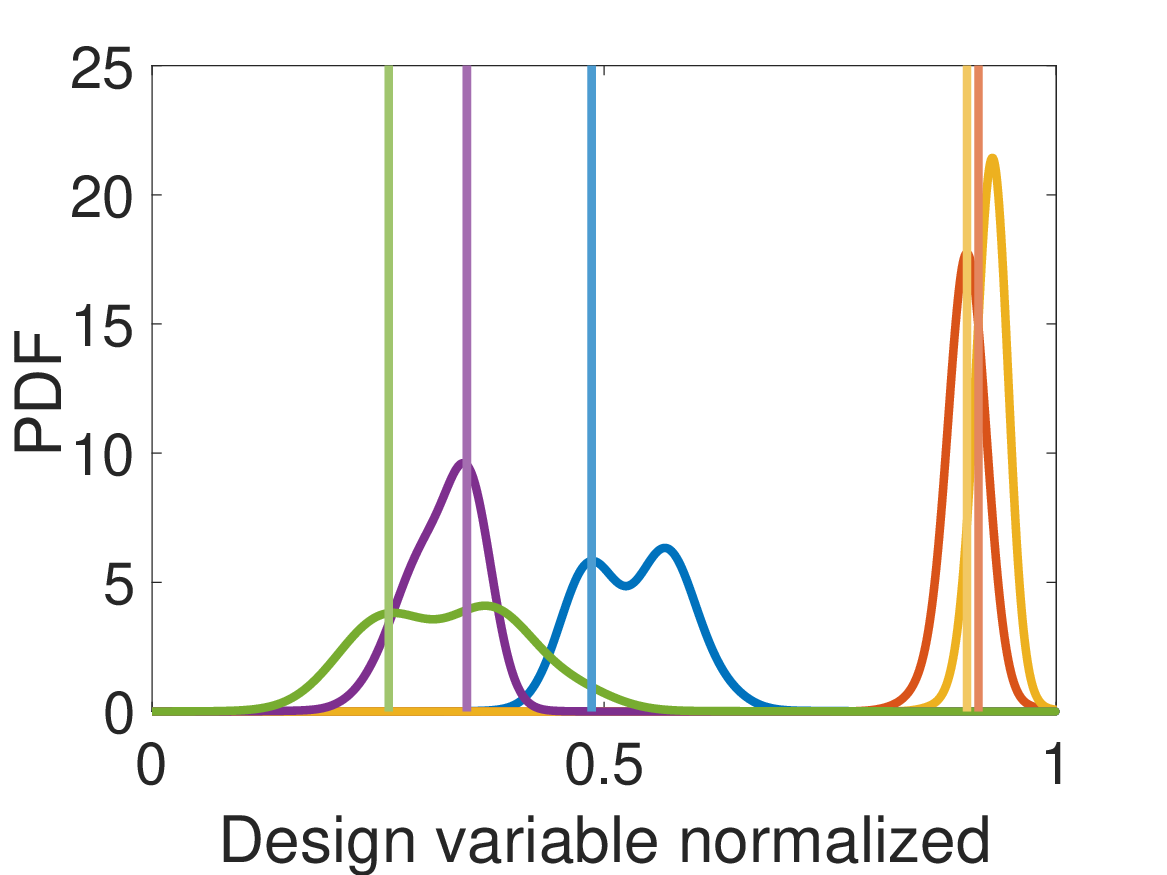}
   \caption{Sum of the pdfs of the individual design parameters weighted by the mixing coefficients for the MDN (top) and MDN+TL1 (bottom) models.}
    \label{fig:results3}
\end{figure}

We now illustrate additional results where $K=10$ for one sample of the test dataset in Figs. \ref{fig:results1}-\ref{fig:results2_2}. For ease of visualization, we focus on the most significant $4$ solutions provided by the MDN without autoencoder and without transfer learning and the MDN without autoencoder and using the first transfer learning strategy. The vertical lines in Fig. \ref{fig:results1_2} and Fig. \ref{fig:results2_2} represent the exact values of the design parameters corresponding to the optical spectrum of the test sample. Looking at the absorbance plot, the blue line represents the test optical spectrum, while the red dashed line represents the spectrum computed by the EM solver using the optimal design parameters values predicted by the MDN. There exist multiple approaches for sampling the MDN output. For the results shown here we have taken the mean values of the individual Gaussian pdfs associated with a mixing coefficient. This is an approach but it is not the only one. Overall, MDNs provide pdfs information that can be used to explore design solutions and as an initial solution generator for further optimization techniques. It can be noted that the MDN with transfer learning is more accurate than without transfer learning, and can well identify multiple design solutions. Fig. \ref{fig:results3} shows the sum of the pdfs of the individual design parameters weighted by the mixing coefficients for the two MDNs compared previously. Similarly, the vertical lines represent the exact values of the design parameters corresponding to the optical spectrum of the test sample. 



\section{Conclusion}

In this work, we have presented a transfer-learning assisted
MDN methodology for inverse modeling. This allows
a rapid exploration of MDNs with different number of
Gaussian distributions, while obtaining accurate results in the
prediction of the design solutions given an optical response
as an input. Numerical results have validated the proposed
techniques.

\section*{Acknowledgement}
F. Ferranti acknowledges support from the Methusalem and
Hercules foundations, and the OZR of the Vrije Universiteit Brussel (VUB). L. Cheng and P. Singh acknowledge resources provided for computations by the National Academic Infrastructure for Supercomputing in Sweden (NAISS), partially funded by the Swedish Research Council through grant agreement no. 2022-06725.


\bibliographystyle{unsrt}  
\bibliography{bibl_file}

\begin{thebibliography}{10}

\bibitem{Ma2018}
Wei Ma, Feng Cheng, and Yongmin Liu.
\newblock Deep-learning-enabled on-demand design of chiral metamaterials.
\newblock {\em {ACS} Nano}, 12(6):6326--6334, June 2018.

\bibitem{Peurifoy2018}
John Peurifoy, Yichen Shen, Li~Jing, Yi~Yang, Fidel Cano-Renteria, Brendan~G.
  DeLacy, John~D. Joannopoulos, Max Tegmark, and Marin Solja{\v{c}}i{\'{c}}.
\newblock Nanophotonic particle simulation and inverse design using artificial
  neural networks.
\newblock {\em Science Advances}, 4(6):eaar4206, June 2018.

\bibitem{Liu2018}
Dianjing Liu, Yixuan Tan, Erfan Khoram, and Zongfu Yu.
\newblock Training deep neural networks for the inverse design of nanophotonic
  structures.
\newblock {\em {ACS} Photonics}, 5(4):1365--1369, February 2018.

\bibitem{Zheng20}
Rohit Unni, Kan Yao, and Yuebing Zheng.
\newblock Deep convolutional mixture density network for inverse design of
  layered photonic structures.
\newblock {\em ACS Photonics}, 7(10):2703--2712, 2020.

\bibitem{Wiecha20}
Peter~R. Wiecha and Otto~L. Muskens.
\newblock Deep learning meets nanophotonics: A generalized accurate predictor
  for near fields and far fields of arbitrary 3d nanostructures.
\newblock {\em Nano Letters}, 20(1):329--338, 2020.

\bibitem{Wiecha21}
Peter~R. Wiecha, Arnaud Arbouet, Christian Girard, and Otto~L. Muskens.
\newblock Deep learning in nano-photonics: inverse design and beyond.
\newblock {\em Photon. Res.}, 9(5):B182--B200, May 2021.

\bibitem{So2021}
Sunae So, Younghwan Yang, Taejun Lee, and Junsuk Rho.
\newblock On-demand design of spectrally sensitive multiband absorbers using an
  artificial neural network.
\newblock {\em Photon. Res.}, 9(4):B153--B158, Apr 2021.

\bibitem{Roberts21}
Nathan~Bryn Roberts and Mehdi Keshavarz~Hedayati.
\newblock {A deep learning approach to the forward prediction and inverse
  design of plasmonic metasurface structural color}.
\newblock {\em Applied Physics Letters}, 119(6):061101, 08 2021.

\bibitem{Unni2021}
Rohit Unni, Kan Yao, Xizewen Han, Mingyuan Zhou, and Yuebing Zheng.
\newblock A mixture-density-based tandem optimization network for on-demand
  inverse design of thin-film high reflectors.
\newblock {\em Nanophotonics}, 10(16):4057--4065, October 2021.

\bibitem{Kojima2021}
Keisuke Kojima, Mohammad~H. Tahersima, Toshiaki Koike-Akino, Devesh~K. Jha,
  Yingheng Tang, Ye~Wang, and Kieran Parsons.
\newblock Deep neural networks for inverse design of nanophotonic devices.
\newblock {\em Journal of Lightwave Technology}, 39(4):1010--1019, 2021.

\bibitem{Xu2021}
Xiaopeng Xu, Chonglei Sun, Yu~Li, Jia Zhao, Junlei Han, and Weiping Huang.
\newblock An improved tandem neural network for the inverse design of
  nanophotonics devices.
\newblock {\em Optics Communications}, 481:126513, 2021.

\bibitem{Wiecha22}
Clément Majorel, Christian Girard, Arnaud Arbouet, Otto~L. Muskens, and
  Peter~R. Wiecha.
\newblock Deep learning enabled strategies for modeling of complex aperiodic
  plasmonic metasurfaces of arbitrary size.
\newblock {\em ACS Photonics}, 9(2):575--585, 2022.

\bibitem{Ferranti22_2}
Francesco Ferranti.
\newblock Feature-based machine learning for the efficient design of
  nanophotonic structures.
\newblock {\em Photonics and Nanostructures - Fundamentals and Applications},
  52:101077, 2022.

\bibitem{Padilla22}
Yang Deng, Simiao Ren, Jordan Malof, and Willie~J. Padilla.
\newblock Deep inverse photonic design: A tutorial.
\newblock {\em Photonics and Nanostructures - Fundamentals and Applications},
  52:101070, 2022.

\bibitem{Yesilyurt23}
Omer Yesilyurt, Samuel Peana, Vahagn Mkhitaryan, Karthik Pagadala, Vladimir~M.
  Shalaev, Alexander~V. Kildishev, and Alexandra Boltasseva.
\newblock Fabrication-conscious neural network based inverse design of
  single-material variable-index multilayer films.
\newblock {\em Nanophotonics}, 12(5):993--1006, 2023.

\bibitem{Luce2023}
Alexander Luce, Ali Mahdavi, Heribert Wankerl, and Florian Marquardt.
\newblock Investigation of inverse design of multilayer thin-films with
  conditional invertible neural networks.
\newblock {\em Machine Learning: Science and Technology}, 4(1):015014, February
  2023.

\bibitem{She22}
Wei She, Renzhong Zhang, Wei Liu, Lihong Zhong, Bin Chen, and Zhao Tian.
\newblock Mixture density network based on truncated distribution and genetic
  algorithm for wind power forecasting.
\newblock {\em Journal of Physics: Conference Series}, 2409, 2022.

\bibitem{Brandimarte14}
P.~Brandimarte.
\newblock {\em Low-Discrepancy Sequences}, pages 379--401.
\newblock {John Wiley and Sons, Inc.}, 2014.

\end{thebibliography}

\end{document}